\documentclass[letterpaper]{article} 
\usepackage{aaai2026}  
\usepackage{times}  
\usepackage{helvet}  
\usepackage{courier}  
\usepackage[hyphens]{url}  
\usepackage{graphicx} 
\usepackage{natbib}  
\usepackage{caption} 
\usepackage[utf8]{inputenc} 
\usepackage[T1]{fontenc}    
\usepackage{url}            
\usepackage{booktabs}       
\usepackage{amsfonts}       
\usepackage{nicefrac}       
\usepackage{microtype}      
\usepackage{xcolor}         
\usepackage{algorithmic}
\usepackage{algorithm}
\usepackage{subfigure}
\usepackage{threeparttable}
\usepackage{graphicx}
\usepackage{subcaption}
\usepackage{amsmath}
\usepackage{array}      
\usepackage{caption}    
\usepackage{algorithm}
\usepackage{algorithmic}
\usepackage{dsfont}
\usepackage{newfloat}
\usepackage{listings}
\DeclareCaptionStyle{ruled}{labelfont=normalfont,labelsep=colon,strut=off} 
\floatstyle{ruled}
\newfloat{listing}{tb}{lst}{}
\floatname{listing}{Listing}
\nocopyright  

\title{Self-Enhanced Image Clustering with Cross-Modal Semantic Consistency}

\title{Self-Enhanced Image Clustering with Cross-Modal Semantic Consistency}
\author {
    Zihan Li\textsuperscript{\rm 1},
    Wei Sun\textsuperscript{\rm 1},
    Jing Hu\textsuperscript{\rm 2},
    Jianhua Yin\textsuperscript{\rm 1},
    Jianlong Wu\textsuperscript{\rm 3},
    Liqiang Nie\textsuperscript{\rm 3}
}
\affiliations {
    \textsuperscript{\rm 1}Shandong University,
    \textsuperscript{\rm 2}Tongji University,
    \textsuperscript{\rm 2}Harhin lnstitute of Technology, Shenzhen\\
    zihanli@mail.sdu.edu.cn, sunweimy@gmail.com, jinghu@utexas, jhyin@sdu.edu.cn, wujianlong@hit.edu.cn, nieliqiang@gmail.com
    }

\usepackage{bibentry}
\begin{document}

\maketitle

\begin{abstract}
While large language-image pre-trained models like CLIP offer powerful generic features for image clustering, existing methods typically freeze the encoder. This creates a fundamental mismatch between the model's task-agnostic representations and the demands of a specific clustering task, imposing a ceiling on performance. To break this ceiling, we propose a self-enhanced framework based on cross-modal semantic consistency for efficient image clustering. Our framework first builds a strong foundation via Cross-Modal Semantic Consistency and then specializes the encoder through Self-Enhancement.
In the first stage, we focus on Cross-Modal Semantic Consistency. By mining consistency between generated image-text pairs at the instance, cluster assignment, and cluster center levels, we train lightweight clustering heads to align with the rich semantics of the pre-trained model. This alignment process is bolstered by a novel method for generating higher-quality cluster centers and a dynamic balancing regularizer to ensure well-distributed assignments. 
In the second stage, we introduce a Self-Enhanced fine-tuning strategy. The well-aligned model from the first stage acts as a reliable pseudo-label generator. These self-generated supervisory signals are then used to feed back the efficient, joint optimization of the vision encoder and clustering heads, unlocking their full potential.
Extensive experiments on six mainstream datasets show that our method outperforms existing deep clustering methods by significant margins. Notably, our ViT-B/32 model already matches or even surpasses the accuracy of state-of-the-art methods built upon the far larger ViT-L/14.
\end{abstract}

\section{Introduction}

Image clustering is a classic unsupervised task that aims to group unlabeled samples into different clusters by exploiting the inherent relationships between the samples. Initially, researchers mainly focused on traditional machine learning methods for image clustering, including K-means~\cite{neyman1967berkeley}, hierarchical clustering~\cite{ward1963hierarchical}, spectral clustering~\cite{ng2001spectral,zelnik2004self}, subspace clustering~\cite{kailing2004density,liu2019robust,zhang2015low,zhang2018generalized}, and concept factorization~\cite{zhang2019flexible,cai2009locality}, most of which relied heavily on prior assumptions about the data distribution. However, their performance decreases significantly when faced with complex and high-dimensional image data. With the emergence of deep learning, researchers began to incorporate deep learning into clustering objectives. Based on the powerful feature extraction capabilities,
deep learning image clustering approaches can automatically and accurately capture the structure and semantics within the data, exhibiting outstanding clustering performance~\cite{tsai2020mice,vincent2010stacked,radford2015unsupervised,zeiler2010deconvolutional,kingma2013auto,yang2016joint,xie2016unsupervised,chang2017deep,wu2019deep,qian2023stable,metaxas2023divclust}.
\begin{figure}[!t]
\centering

\includegraphics[width=0.48\textwidth]
{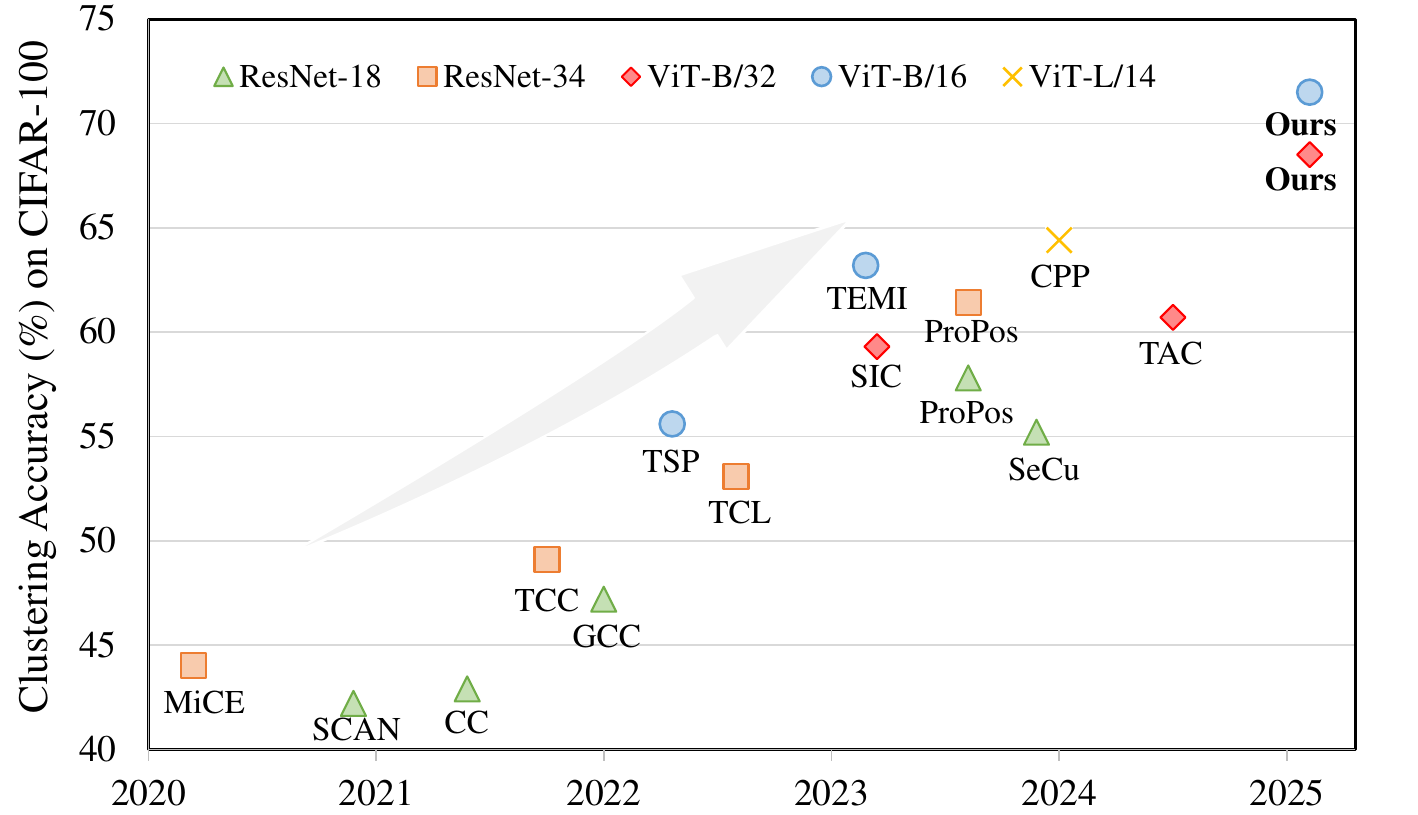}
\caption{Comparison of existing deep clustering methods. The deep clustering paradigm gradually shifts from CNN-based to ViT-based with a steady increase in performance.
}

\label{figIntro}
\end{figure}
With the development of Vision Transformers~(ViT)~\cite{dosovitskiy2020image}, the paradigm of deep clustering is gradually shifting from traditional CNN-based methods to ViT-based approaches, as illustrated in Fig.~\ref{figIntro}. Compared with convolutional neural networks that capture local features through sliding convolutional kernels, ViTs employ a self-attention mechanism, which can effectively capture the global semantic information, thus obtaining higher-quality features and improving clustering results. Due to the huge amount of parameters in ViT, training ViT from scratch in an end-to-end manner requires significant computational resources. As a result, existing ViT-based clustering methods, such as TSP~\cite{zhou2022deep}, TEMI~\cite{adaloglouexploring}, CPP~\cite{chu2023image}, etc., typically rely on pre-trained vision encoders that are completely frozen during training and introduce a trainable projection head and clustering head to simultaneously learn task-specific features and cluster assignments. Considering the limitation that relying solely on visual features for training may incorrectly cluster visually similar but semantically dissimilar samples together, recent methods like SIC~\cite{cai2023semantic} and TAC~\cite{li2023image} have further incorporated additional textual information from the WordNet~\cite{miller1995wordnet} dictionary to assist in image clustering tasks, successfully enhancing clustering performance. However, this frozen-encoder approach introduces a fundamental bottleneck: it creates a mismatch between the general-purpose, task-agnostic features of the pre-trained model and the highly specific, discriminative representations needed for the target clustering task. This inevitably caps the model's performance, regardless of how well the attached clustering heads are trained. The key, therefore, is to find a way to safely unlock and adapt the encoder to the specific task without supervised labels.

To address this limitation, we propose a novel two-stage framework, \textbf{S}elf-\textbf{E}nhanced \textbf{I}mage \textbf{C}lustering (SEIC), designed to first align with the pre-trained knowledge and then specialize the encoder for the task at hand. Our first step is to build a strong foundational model without altering the powerful pre-trained encoder. We start by generating image-text feature pairs using knowledge from CLIP and WordNet. Then, we train the clustering heads by enforcing Cross-Modal Semantic Consistency at three distinct levels: instance features, cluster assignments, and cluster centers. This multi-level alignment forces the heads to learn representations that are consistent with the rich semantics of the CLIP space. To strengthen this alignment, we introduce a novel method for computing higher-quality cluster centers based on assignment probabilities and a dynamic balance regularizer that adapts to the sample learning status to mitigate cluster collapse.Once the heads are well-aligned, the model can produce reliable, high-confidence pseudo-labels. In the second stage, we leverage this capability for Self-Enhancement. The high-quality pseudo-labels serve as self-generated supervisory signals that feed back to guide the fine-tuning process. Using efficient parallel LoRA adapters~\cite{hu2021lora}, we jointly optimize the vision encoder and the clustering heads. This critical step allows the encoder to move beyond its generic initialization and specialize its feature extraction capabilities for the specific data distribution, thereby breaking the performance ceiling imposed by the frozen-encoder paradigm.

Overall, our contributions can be summarized as follows:
\begin{itemize}
    \item We propose a novel two-stage \textbf{"align, then enhance"} framework (SEIC) that resolves the feature mismatch problem in pre-trained models by first building a strong foundational model and then specializing the encoder for clustering.
    \item For the alignment stage, we enforce Cross-Modal Semantic Consistency at multiple levels, supported by a novel probability-weighted cluster center generation method and a dynamic balancing regularizer.
    \item For the enhancement stage, we introduce a Self-Enhanced Efficient Fine-tuning where the model uses its own high-confidence predictions to feed back and efficiently refine the vision encoder via LoRA.
    \item Extensive experiments on six benchmark datasets demonstrate that SEIC significantly outperforms existing state-of-the-art methods, even when using smaller backbones.
\end{itemize}

\section{Related work}
\subsection{Pre-trained ViT Models}
The success of deep clustering is increasingly tied to powerful pre-trained models. Self-supervised methods like MoCo-v3~\cite{chen2021empirical}, MAE~\cite{he2022masked}, and DINO~\cite{caron2021emerging} learn robust visual representations from images alone. More recently, vision-language models such as CLIP~\cite{radford2021learning} have shown remarkable zero-shot capabilities by learning from massive image-text pairs. Our work investigates how to effectively adapt the rich knowledge from these models, particularly CLIP, for the unsupervised clustering task.


However, the research on utilizing the knowledge of pre-trained models to improve the image clustering task is still in the exploratory stage. Therefore, in this paper, we propose a self-enhanced framework based on the pre-trained model with both image and text modalities to provide more insights and solutions for this issue.

\begin{figure*}[!t]
\centering
\includegraphics[width=0.95\textwidth]{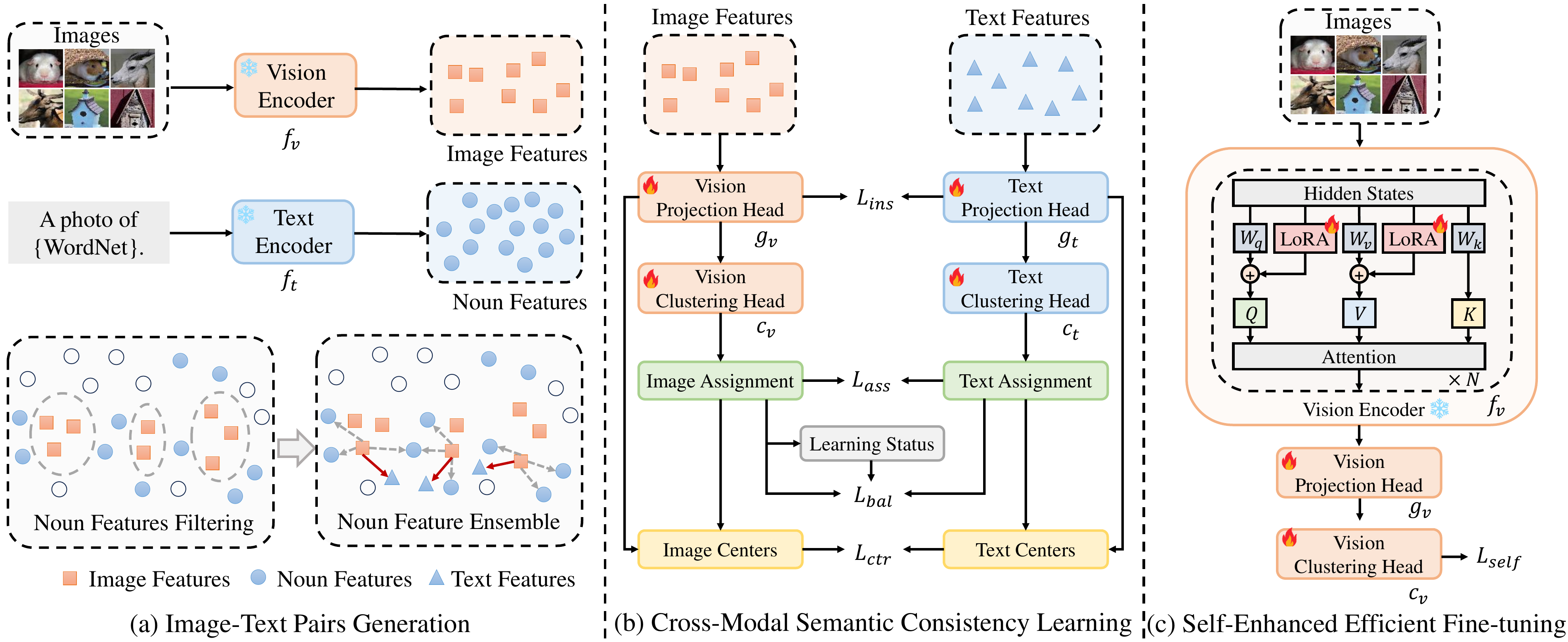}
\caption{The framework of our proposed SEIC method. (a) We first construct image-text feature pairs using the pre-trained model CLIP and the dictionary WordNet. (b) Then, we perform cross-modal semantic consistency learning to achieve more discriminative representations and accurate clustering assignments. (c) Finally, a self-enhanced fine-tuning strategy is proposed to further improve the clustering performance.}
\label{framework}
\end{figure*}

\subsection{Deep Image Clustering}
In the past, deep image clustering has primarily relied on ResNet~\cite{he2016deep} as the backbone, enhancing its performance through various optimization techniques. Recently, the introduction of contrastive learning has significantly advanced deep image clustering, with methods such as ~\cite{li2021contrastive,zhong2021graph,shen2021you,li2022twin,van2020scan,li2020prototypical,huang2022learning,niu2022spice,metaxas2023divclust,qian2023stable}. Among them, CC~\cite{li2021contrastive}, GCC~\cite{zhong2021graph}, and TCL~\cite{li2022twin} utilize contrastive learning at both the instance and cluster levels.  CC and TCL create positive and negative samples through data augmentation, while GCC expands the selection of such samples through K-nearest neighbors.  PCL~\cite{li2020prototypical} and ProPos~\cite{huang2022learning} focus solely on instance comparison, utilizing prototype contrast to enrich the selection of positive samples.  Some methods based on pseudo-label have also been proposed to further improve clustering model performance, such as SCAN~\cite{van2020scan} and SPICE~\cite{niu2022spice}.  SCAN introduces the self-labeling method, further enhancing model performance by selecting high-confidence pseudo-labels from a trained clustering model. SPICE employs semi-supervised training by fixing a portion of pseudo-labels for model fine-tuning.

Recently, several deep image clustering methods have sought to advance clustering tasks by utilizing pre-trained ViT models.  TEMI~\cite{adaloglouexploring} employs DINO~\cite{caron2021emerging} to reduce the distance between images and their neighbors.  TSP~\cite{zhou2022deep} undergoes training through self-distillation and is trained on the necessary pre-trained model.  SIC~\cite{cai2023semantic} and TAC~\cite{li2023image} predominantly use the CLIP~\cite{radford2021learning} pre-trained model, effectively leveraging its characteristics by constructing text features.  However, all of these models freeze the encoders and train only the appended heads, limiting the model to task-agnostic features. In our approach, we learn discriminative representations and correct assignments by mining multi-level cross-modal semantic consistency and propose a self-enhanced fine-tuning strategy to jointly fine-tune encoder and head components.

\subsection{Parameter-Efficient Fine-Tuning}
Fine-tuning large models is computationally expensive. PEFT methods, such as adapters~\cite{houlsby2019parameter} and Low-Rank Adaptation (LoRA)~\cite{hu2021lora}, address this by updating only a small number of parameters. While these techniques are well-established in supervised learning, their application in a fully unsupervised clustering scenario remains less explored. Our work demonstrates an effective way to leverage LoRA for unsupervised encoder specialization.

\section{Method}
\label{sec:2_method}

\subsection{Problem Formulation and Overview of SEIC}
Given $N$ unlabeled images $\mathcal{I} =[I_1, ..., I_N]$, deep clustering aims to group them into $K$ different clusters, ensuring that images of the same semantic labels are grouped together as much as possible. As shown in Fig.~\ref{framework},  we design a self-enhanced framework SEIC for image clustering based on image-text pairs with the help of the pre-trained model CLIP~\cite{radford2021learning}. The model consists of vision and text pre-trained encoders and corresponding head components. Specifically, the vision encoder $f_v$ and text encoder $f_t$ are able to transform the images as well as the corresponding texts into features, leveraging the knowledge of CLIP. The vision projection head $g_v$ and text projection head $g_t$ aim to project features into a space more suitable for clustering. The clustering heads $c_v$ and $c_t$ are used for cluster assignment, i.e., mapping the projected image or text feature to a $K$-dimensional probability vector $p_i$ that satisfies $\sum_{j=1}^{K}p_{ij}=1$, where $p_{ij}$ denotes the probability that the image $I_i$ is assigned to the $j$-th cluster. Finally, $r_i=\arg \max (p_i)$ will be treated as the clustering result for image $I_i$.

The training process of our framework can be divided into two stages: 

\noindent1) \textbf{Alignment via Cross-Modal Semantic Consistency}: We first freeze the CLIP encoders and train only the head components. The goal is to align the heads' behavior with the rich, latent semantic space of CLIP, thereby creating a strong foundational clustering model.

\noindent2) \textbf{Self-Enhancement for Encoder Specialization}: With a well-aligned model, we then proceed to fine-tune the vision encoder $f_v$ itself. The model uses its own high-confidence predictions to generate supervisory signals, which \textit{feed back} to specialize the encoder for the target dataset.

\subsection{Alignment via Cross-Modal Semantic Consistency}
In this stage, our objective is to learn high-quality projection and clustering heads by maximally extracting knowledge from the frozen CLIP encoders. This process involves generating cross-modal data and then enforcing consistency at multiple semantic levels.

\subsubsection{Image-Text Pair Generation}

In the deep clustering task, where image samples lack textual annotations, our goal is to construct image-text pairs $\mathcal{S}=\{v_i,t_i\}_{i=1}^{N}$. Based on the premise that nouns closer to an image in the shared feature space better reflect its content, we utilize the WordNet dictionary to generate textual information for each image.

Similar to TAC~\cite{li2023image}, specifically, we first extract image features $\mathcal{V}=\{v_i\}_{i=1}^{N}$ using the frozen vision encoder $f_v$. All nouns from WordNet are encoded into features $\mathcal{N}$ using the text encoder $f_t$. To manage computational costs, we create a candidate noun subset $\overline{\mathcal{N}}$ by selecting the top $k_1$ nearest noun features for each of the $K$ initial image cluster centers (obtained via K-means). Finally, for each image feature $v_i$, its corresponding text feature $t_i$ is constructed as a weighted sum of its top-$k_2$ most similar noun features from $\overline{\mathcal{N}}$:
\begin{equation}
\label{l_tf}
t_i=\sum_{j=1}^{|\overline{\mathcal{N}}|}\hat{s}_{ij}\overline{n}_j,
\end{equation}
where $\hat{s}_{ij}$ is non-zero only if $\overline{n}_j$ is one of the top-$k_2$ neighbors of $v_i$. This provides us with the image-text pairs $\{v_i, t_i\}$ needed for alignment.

\subsubsection{Cross-Modal~Semantic~Consistency~Learning}
With the image-text pairs, we now align the heads by enforcing consistency at the instance feature, cluster assignment, and cluster center levels. 

For instance feature level consistency, we treat the image and text projected features obtained from the projection heads as positive pairs for contrastive learning. To be specific, we first compute the projected features $\widetilde{v}_i=g_v(v_i),\widetilde{t}_i=g_t(t_i)$, based on which we construct the bidirectional contrastive loss as follows:
\begin{equation}
L_{ins}^{(v\to t)}=-\frac{1}{B}\sum_{i=1}^B \log \frac{\exp(\langle \widetilde{v}_i,\widetilde{t}_i \rangle/\tau)}{\sum_{j=1}^B \exp(\langle \widetilde{v}_i,\widetilde{t}_j \rangle/\tau)},
\end{equation}
\begin{equation}
\label{eq_cross}
L_{ins}=(L_{ins}^{(v\to t)}+L_{ins}^{(t\to v)})/2,
\end{equation}
where $\tau$ is a trainable scaling factor and $B$ is the batch size. $L_{ins}$ is structurally similar to the loss proposed in CLIP, which allows the projection heads to effectively align with the distribution of image-text features learned by CLIP, thereby generating more clustering-friendly features.

At the cluster assignment level, the clustering results from the image and text clustering heads should also share consistent distributions. Therefore, we first achieve the cluster assignment probabilities under different modalities based on the projected feature, i.e.,  $p^v_i=c_v(\widetilde{v}_i)$, $p^t_i=c_t(\widetilde{t}_i)$. For each column of $p^v$ and $p^t$ within a mini-batch, we treat them as a positive pair and perform contrastive learning as follows: 
\begin{equation}
L_{ass}^{(v\to t)}=-\frac{1}{K}\sum_{i=1}^K \log\frac{\exp(\langle \ddot{p^v_i},\ddot{p^t_i} \rangle/\hat{\tau})}{\sum_{j=1}^K \exp(\langle \ddot{p^v_i},\ddot{p^t_j} \rangle/\hat{\tau})},
\end{equation}
\begin{equation}
\label{eq_assign}
L_{ass}=(L_{ass}^{(v\to t)}+L_{ass}^{(t\to v)})/2,
\end{equation}
where $\ddot{p^v_i}$ represents the $i$-th column of $p^v$, and $\hat{\tau}$ is a fixed temperature coefficient. This loss encourages the clustering heads to produce more accurate cluster assignments.

\begin{table*}[!t]
\renewcommand\arraystretch{1.1}
\caption{Clustering results of various methods on six widely-used datasets. The best and second-best results are shown in \textbf{bold} and \underline{underline}, respectively.}
\label{mainresultTab}
\resizebox{\linewidth}{!}{
\begin{tabular}{cccccccccccccccccccc}
\hline
\multicolumn{2}{c}{Dataset}            & \multicolumn{3}{c}{CIFAR-10}                                                & \multicolumn{3}{c}{CIFAR-100}                                               & \multicolumn{3}{c}{STL-10}                                                  & \multicolumn{3}{c}{ImageNet-10}                                              & \multicolumn{3}{c}{ImageNet-Dogs}                                           & \multicolumn{3}{c}{Tiny-ImageNet}                                           \\ \hline
\multicolumn{1}{c}{Method} & Backbone  & \multicolumn{1}{c}{NMI} & \multicolumn{1}{c}{ACC} & \multicolumn{1}{c}{ARI} & \multicolumn{1}{c}{NMI} & \multicolumn{1}{c}{ACC} & \multicolumn{1}{c}{ARI} & \multicolumn{1}{c}{NMI} & \multicolumn{1}{c}{ACC} & \multicolumn{1}{c}{ARI} & \multicolumn{1}{c}{NMI} & \multicolumn{1}{c}{ACC} & \multicolumn{1}{c}{ARI} & \multicolumn{1}{c}{NMI} & \multicolumn{1}{c}{ACC} & \multicolumn{1}{c}{ARI} & \multicolumn{1}{c}{NMI} & \multicolumn{1}{c}{ACC} & \multicolumn{1}{c}{ARI} \\ \hline
K-means                    & -         & 0.087                   & 0.229                   & 0.049                   & 0.084                   & 0.130                   & 0.028                   & 0.125                   & 0.192                   & 0.061                   & 0.119                   & 0.241                   & 0.057                   & 0.055                   & 0.105                   & 0.020                   & 0.065                   & 0.025                   & 0.005                   \\
SCAN                       & ResNet-18 & 0.797                   & 0.883                   & 0.772                   & 0.486                   & 0.507                   & 0.333                   & 0.698                   & 0.809                   & 0.646                   & -                       & -                       & -                       & -                       & -                       & -                       & -                       & -                       & -                       \\
GCC                        & ResNet-18 & 0.764                   & 0.856                   & 0.728                   & 0.472                   & 0.472                   & 0.305                   & 0.684                   & 0.788                   & 0.631                   & 0.842                   & 0.901                   & 0.822                   & 0.490                   & 0.526                   & 0.322                   & 0.347                   & 0.138                   & 0.075                   \\
ProPos                     & ResNet-18 & 0.853                   & 0.921                   & 0.844                   & 0.572                   & 0.573                   & 0.417                   & -                       & -                       & -                       & -                       & -                       & -                       & -                       & -                       & -                       & 0.405                   & 0.256                   & 0.143                   \\
ProPos                     & ResNet-34 & 0.886                   & 0.943                   & 0.884                   & 0.606                   & 0.614                   & 0.451                   & 0.758                   & 0.867                   & 0.737                   & 0.896                   & 0.956                   & 0.906                   & 0.692                   & 0.745                   & 0.627                   & -                       & -                       & -                       \\
TCL                        & ResNet-34 & 0.819                   & 0.887                   & 0.780                   & 0.529                   & 0.531                   & 0.357                   & 0.799                   & 0.868                   & 0.757                   & 0.875                   & 0.895                   & 0.837                   & 0.623                   & 0.644                   & 0.516                   & -                       & -                       & -                       \\
TCC                        & ResNet-34 & 0.790                    & 0.906                   & 0.733                   & 0.479                   & 0.491                   & 0.312                   & 0.732                   & 0.814                   & 0.689                   & 0.848                   & 0.897                   & 0.825                   & 0.554                   & 0.595                   & 0.417                   & -                       & -                       & -                       \\
TSP                        & DINO ViT-S/16  & 0.847                   & 0.921                   & 0.838                   & 0.582                   & 0.549                   & 0.408                   & 0.941                   & 0.970                    & 0.938                   & -                       & -                       & -                       & -                       & -                       & -                       & -                       & -                       & -                       \\
TSP                        & DINO ViT-B/16  & 0.880                    & 0.940                    & 0.875                   & 0.614                   & 0.556                   & 0.433                   & 0.958                   & 0.979                   & 0.956                   & -                       & -                       & -                       & -                       & -                       & -                       & -                       & -                       & -                       \\
SIC                        & CLIP ViT-B/32  & 0.847                   & 0.926                   & 0.844                   & 0.593                   & 0.583                   & 0.439                   & 0.953                   & 0.981                   & 0.959                   & 0.970                    & 0.982                   & 0.961                   & 0.690                    & 0.697                   & 0.558                   & -                       & -                       & -                       \\
TEMI                       & DINO ViT-B/16  & 0.886                   & 0.945                   & 0.885                   & 0.654                   & 0.632                   & 0.489                   & \underline{0.965}                   & 0.985                   & 0.968                   & -                       & -                       & -                       & -                       & -                       & -                       & -                       & -                       & -                       \\
TEMI                       & CLIP ViT-L/14  & 0.926                   & 0.969                   & 0.932                   & 0.645                   & 0.618                   & 0.468                   & 0.964                   & 0.974                   & 0.949                   & -                       & -                       & -                       & -                       & -                       & -                       & -                       & -                       & -                       \\
CPP                        & CLIP ViT-L/14  & 0.936                   & 0.974                   & \multicolumn{1}{c}{-}   & \underline{0.725}                   & 0.642                   & \multicolumn{1}{c}{-}   & -                       & -                       & -                       & -                       & -                       & -                       & -                       & -                       & -                       & -                       & -                       & -                       \\
TAC                        & CLIP ViT-B/32  & 0.841                   & 0.923                   & 0.839                   & 0.611                   & 0.607                   & 0.448                   & 0.955                   & 0.982                   & 0.961                   & 0.985                   & 0.992                   & 0.983                   & 0.806                   & 0.830                    & 0.722                   & -                       & -                       & -                       \\

SEIC (Ours)                       & CLIP ViT-B/32  & \underline{0.938}                   & \underline{0.976}                   & \underline{0.947}                   & 0.691                   & \underline{0.685}                   & \underline{0.554}                   & 0.963    & \underline{0.987}    & \underline{0.970}    & \underline{0.988}    & \underline{0.996}    & \underline{0.991}    & \underline{0.846}    & \underline{0.889}    & \underline{0.817}    & \underline{0.701}    & \underline{0.598}    & \underline{0.472}    \\


SEIC (Ours)                       & CLIP ViT-B/16  & \textbf{0.945}                   &    \textbf{0.978}                & \textbf{0.952}                   & \textbf{0.732}                   & \textbf{0.715}                   & \textbf{0.598}                   & \textbf{0.980}                    & \textbf{0.993}                   & \textbf{0.985}                   & \textbf{0.992}                   & \textbf{0.997}                   & \textbf{0.994}                   & \textbf{0.886}                   & \textbf{0.910}                    & \textbf{0.858}                   & \textbf{0.730}                    & \textbf{0.644}                   & \textbf{0.523}                   \\ \hline
\end{tabular}}

        

\end{table*}

Considering that we only explore feature consistency at the instance level, which ignores potential category information between samples, we further extend our approach to incorporate cluster center-level consistency. It ensures that the centers sharing the same category across different modalities are maximally proximate in feature space, while those of different categories are significantly distanced apart.
To achieve more accurate cluster centers, we propose a novel generation method. Unlike the conventional approach of directly taking the mean of all sample features within a cluster as the center~\cite{li2020prototypical,huang2022learning}, we construct the center according to the contribution of samples to this cluster, which is reflected by the clustering predicted probability. 
Specifically, each cluster center is formed by the weighted sum of projection features from samples belonging to this center based on their contribution degree, calculated as follows:
\begin{equation}
\mu^v_k=\sum_{i=1}^B 
 \mathds{1}({\mathop{\arg\max}(p^v_{i})}=k) \cdot \bar{p}^v_{ik}\cdot\widetilde{v}_i,
\end{equation}
\begin{equation}
\mu^t_k=\sum_{i=1}^B \mathds{1}({\mathop{\arg\max}(p^t_{i})}=k) \cdot  \bar{p}^t_{ik}\cdot\widetilde{t}_i,
\end{equation}
where $\mathds{1}$ is the indicator function and $\bar{p}_{i}$ is the L1-normalized version of the probability vector $p_{i}$. We then treat the cluster centers of the same category under the image and text modalities as a positive pair and bring them closer in the feature space:
\begin{equation}
\label{eq_center}
L_{ctr}^{(v\to t)}=-\frac{1}{K}\sum_{i=1}^K \log\frac{\exp(\langle \mu^v_i,\mu^t_i \rangle/\hat{\tau})}{\sum_{j=1}^K \exp(\langle \mu^v_i,\mu^t_j \rangle/\hat{\tau})}.
\end{equation}
\begin{equation}
\label{eq_center2}
L_{ctr}=(L_{ctr}^{(v\to t)}+L_{ctr}^{(t\to v)})/2.
\end{equation}
Compared with the traditional PCL~\cite{li2020prototypical} and ProPos~\cite{huang2022learning} methods, our proposed construction method can weigh samples within a batch based on their proximity to these centers. We facilitate the emergence of high-quality cluster centers.

\subsubsection{Stabilizing Alignment with Dynamic Balancing}

To prevent the trivial solution of assigning all samples to a few clusters, we introduce a dynamic balancing regularizer. Instead of a static entropy loss, our term adapts to the model's learning state. We track the historical assignment distribution $h$ for the image modality via an exponential moving average (EMA): $h \leftarrow m \cdot h+(1-m)\cdot \text{Hist}(\arg\max(p^v))$. The balance loss is then defined as:
\begin{equation}
\label{eq_balance}
L_{bal}=\sum_{j=1}^K \frac{b^v_j\log(b^v_j)+b^t_j\log(b^t_j)}{h_j},
\end{equation}
where $b^m_j$ is the average predicted probability for cluster $j$ in the current batch for modality $m \in \{v, t\}$. This dynamically up-weights the penalty for under-populated clusters, encouraging a more uniform assignment.

The final objective for the alignment stage is a weighted sum of these components:
\begin{equation}
\label{eq_total}
L_{align}=\alpha L_{ins}+\beta L_{ass}+\gamma L_{ctr}+\delta L_{bal}. 
\end{equation}

\subsection{Self-Enhanced Efficient Fine-tuning}
The alignment stage yields high-quality clustering heads, but the vision encoder remains generic. To break this performance ceiling, we now unfreeze the vision encoder and specialize it for the target dataset.

A naive approach would be to continue training with $L_{align}$. However, this fails because the text features $\{t_i\}$ are static, pre-computed by the frozen text encoder. As the vision encoder $f_v$ updates, its features $v_i$ will drift, breaking the semantic correspondence with the fixed $t_i$ and leading to performance collapse. We need a new supervisory signal that relies only on the visual modality.

The well-aligned model from Stage 1 is now a confident "teacher" capable of generating reliable pseudo-labels for itself. We use these pseudo-labels to create a self-supervision loss. To handle potential noise in the pseudo-labels, we adopt a confidence-based weighting strategy inspired by SoftMatch~\cite{chen2023softmatch}. A sample's contribution to the loss is weighted by its prediction confidence, measured by a truncated Gaussian function:
\begin{equation}
\label{eq_gaussian}
w_i=  \begin{cases}\exp\left(-\frac{(\max(p_i^v)-\mu_{t})^2}{2\sigma_{t}^2}\right),& \text{if } \max(p^v_i) < \mu_{t}, \\ 1, & \text{otherwise},\end{cases}
\end{equation}
where $\mu_t$ and $\sigma_t^2$ are the moving average of the mean and variance of the maximum prediction probabilities, respectively. The self-enhancement loss is then a weighted cross-entropy:
\begin{equation}
\label{eq_upef}
L_{self}=\frac{1}{B}\sum_{i=1}^B w_i\cdot H(q_i^v, \arg\max(p_i^v)),
\end{equation}
where $H$ is the cross-entropy loss, $q_i^v$ is the prediction for an augmented view of image $I_i$, and $\arg\max(p_i^v)$ is the pseudo-label from the original view.

To fine-tune the vision encoder efficiently and prevent catastrophic forgetting, we integrate lightweight Low-Rank Adaptation (LoRA)~\cite{hu2021lora} adapters into the $Q$ and $V$ matrices of its self-attention blocks. The final optimization in this stage jointly trains the LoRA parameters in $f_v$ along with the vision heads ($g_v, c_v$) using $L_{self}$. This self-enhancement process allows the encoder to learn task-specific features, leading to a significant performance boost.

\section{EXPERIMENTS}

\subsection{Compared Methods}

We compared traditional clustering methods and deep image clustering methods using backbones of various sizes. The baseline methods include K-means~\cite{neyman1967berkeley}, SCAN~\cite{van2020scan}, GCC~\cite{zhong2021graph}, ProPos~\cite{huang2022learning}, TCL~\cite{li2022twin}, TCC~\cite{shen2021you}, TSP~\cite{zhou2022deep}, SIC~\cite{cai2023semantic}, TEMI~\cite{adaloglouexploring}, CPP~\cite{chu2023image}, and TAC~\cite{li2023image}, all of which utilize ResNet-18, ResNet-34, ViT-S/16, ViT-B/32, ViT-B/16 and  ViT-L/14 as backbone.

\begin{table}[!t]
\caption{
The results of different backbones. Metric: ACC. $\dagger$ indicates that the self-enhanced efficient fine-tuning strategy is not used.}
\label{mlpresult}\centering
\resizebox{0.9\linewidth}{!}{\begin{tabular}{cccc}
\hline
Method & Backbone & CIFAR-100 & ImageNet-Dogs \\ \hline
SIC    & CLIP ViT-B/32 & 0.583     & 0.697         \\
TAC    & CLIP ViT-B/32 & 0.607     & 0.830          \\
~SEIC (ours)$^\dagger$   & CLIP ViT-B/32 & 0.608     & 0.833         \\
SEIC (ours)   & CLIP ViT-B/32 & 0.685     & 0.889         \\\hline
TAC    & CLIP ViT-B/16 & -         & 0.857         \\
TEMI   & DINO ViT-B/16 & 0.632     & -             \\
~SEIC (ours)$^\dagger$   & CLIP ViT-B/16 & 0.644     & 0.869         \\
SEIC (ours)   & CLIP ViT-B/16 & 0.715     & 0.910         \\ \hline
TEMI   & CLIP ViT-L/14 & 0.618     & -             \\
CPP    & CLIP ViT-L/14 & 0.642     & -             \\
~SEIC (ours)$^\dagger$   & CLIP ViT-L/14 & 0.698     & 0.893         \\ \hline
\end{tabular}}

        
        

\end{table}

\subsection{Main Results}
To validate the effectiveness of SEIC, we conducted extensive comparisons with state-of-the-art methods across all six datasets. The main results are presented in Tab.~\ref{mainresultTab}. Our method consistently outperforms existing approaches across all metrics and backbones. For instance, on CIFAR-100, our ViT-B/16 model achieves an ACC of 71.5\%, surpassing the DINO-based TEMI (63.2\%) and even the larger ViT-L/14-based CPP (64.2\%) by a significant margin. This highlights the effectiveness of our "align, then enhance" strategy.

Tab.~\ref{mlpresult} further details the performance across different backbone sizes, demonstrating the consistent advantage of our two-stage approach. The full SEIC model substantially improves upon its Stage 1 variant (SEIC$^\dagger$), confirming the benefits of the self-enhancement stage. Notably, our SEIC$^\dagger$ (Stage 1 only) is already competitive with, or superior to, strong baselines like TAC. In terms of efficiency, on an RTX 3090, the alignment stage for CIFAR-100 takes only 5 minutes, and the self-enhancement stage takes approximately 2 hours.

\begin{table}[!t]
\caption{
Effectiveness of individual components. Backbone: ViT-B/16. Metric: ACC.}
\label{ablationComponents}\centering
\resizebox{0.8\linewidth}{!}{\begin{tabular}{ccccc}
\hline
$L_{ins}$  & $L_{ass}$  & $L_{ctr}$ & CIFAR-100 & Tiny-ImageNet \\ \hline
             &              &              & 0.082     & 0.030          \\
             &              & $\checkmark$ & 0.403     & 0.463          \\
             & $\checkmark$ &              & 0.604     & 0.024          \\
$\checkmark$ &              &              & 0.190     & 0.138          \\
             & $\checkmark$ & $\checkmark$ & 0.637     & 0.536          \\
$\checkmark$ &              & $\checkmark$ & 0.541     & 0.486          \\
$\checkmark$ & $\checkmark$ &              & 0.631     & 0.287          \\
$\checkmark$ & $\checkmark$ & $\checkmark$ & 0.644     & 0.561          \\ \hline
\end{tabular}}
\end{table}

\subsection{Ablation Study}
In this section, we dissect the contributions of the key components within our framework. Unless otherwise specified, these experiments are conducted using the Stage 1 model to isolate the effects of the alignment components.

\subsubsection{Effectiveness of Individual Components}
As shown in Tab.~\ref{ablationComponents}, we evaluated the effect of different levels of semantic consistency on clustering performance.  It can be observed that the removal of any component~(especially for cluster assignment and the cluster center level)~results in a significant performance decrease. In addition, different datasets exhibit varying gains from different losses. For example, omitting the center-level consistency loss leads to a 28\% performance drop on Tiny-ImageNet while only a 1\% drop on CIFAR-100, which demonstrates that the cluster center level consistency is crucial for performance improvement, particularly in scenarios with a high number of categories.

\begin{table}[!t]
\caption{
Effectiveness of balance regularization term. Backbone: ViT-B/16. Metric: ACC.}
\label{ablationBR}\centering
\resizebox{0.85\linewidth}{!}{\begin{tabular}{cc}
\hline
Method                                           & Tiny-ImageNet \\ \hline
SEIC (ours) w/o balance term                       & 0.390         \\
SEIC (ours) w/ original balance term & 0.490         \\
SEIC (ours)                                            & 0.561         \\ \hline
\end{tabular}}
\end{table}

\begin{table}[!t]
\caption{
Effectiveness of center generation. Backbone: ViT-B/16. Metric: ACC.}
\label{ablationCenterGeneration}\centering
\resizebox{0.75\linewidth}{!}{\begin{tabular}{ccc}
\hline
Method & CIFAR-100 & Tiny-ImageNet \\ \hline
Mean strategy       & 0.561     & 0.438         \\
SEIC (ours)       & 0.644     & 0.561         \\ \hline
\end{tabular}}
\end{table}

\begin{figure}[t]
\centering
\includegraphics[width=0.33\textwidth]{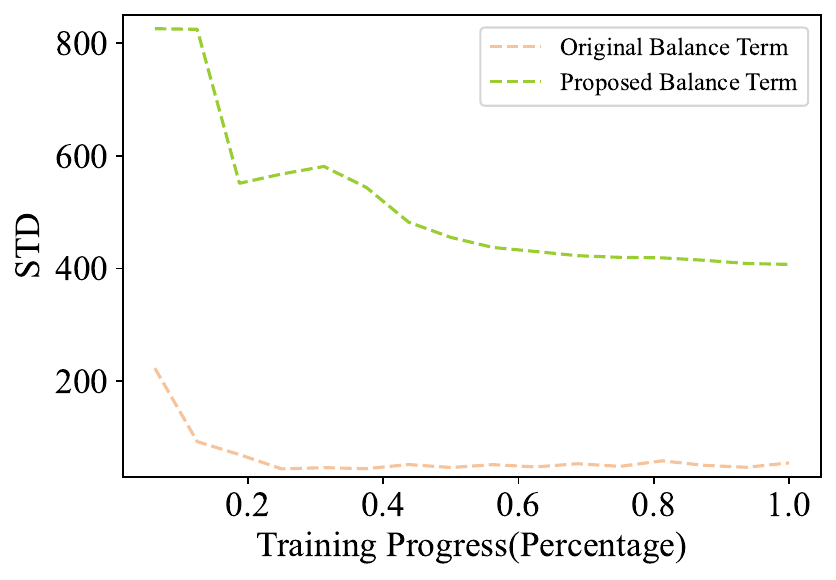}
\caption{Dynamic changes in the standard deviation of the distribution of generated pseudo-labels during the training process. A lower standard deviation indicates a more balanced assignment distribution.} 
\label{figBR}
\end{figure}

\subsubsection{Effectiveness of Balance Regularization Term}
We also explored the effectiveness of our proposed new balance regularization term. The results are presented in Tab.~\ref{ablationBR}. We can see that our designed regularization term can bring 7\% performance improvement compared to the old one. Additionally, as shown in Fig.~\ref{figBR}, we visually illustrated the dynamic changes in the standard deviation of the distribution of generated pseudo-labels during the training process to highlight that our proposed balance term could produce more uniformly distributed pseudo-labels, thus further mitigating the clustering collapse issues.

\subsubsection{Effectiveness of Center Generation}
In order to verify the effectiveness of our designed method of constructing cluster centers, we compared it with the traditional way, i.e., taking the mean of the sample features within the clusters as the center. The comparison results are shown in Tab.~\ref{ablationCenterGeneration}. In contrast to the traditional way, our proposed weight-based construction method shows significant advantages,  especially on Tiny-ImageNet, where the gap reaches 12.3\%, which strongly demonstrates that our method enables the generation of more precise and reasonable centers.




\begin{figure*}[!t]
\centering  

\subfigure[Influence of $k_1$.]{ 
\begin{minipage}{4.7cm}
\flushleft     
\includegraphics[scale=0.35]{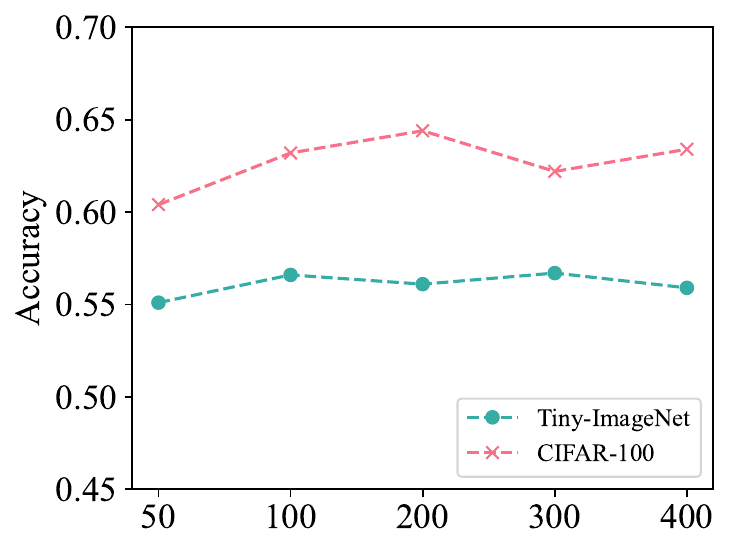}
\end{minipage}
}
\qquad
\subfigure[Influence of $k_2$.]{ 
\begin{minipage}{4.7cm}
\flushleft     
\includegraphics[scale=0.35]{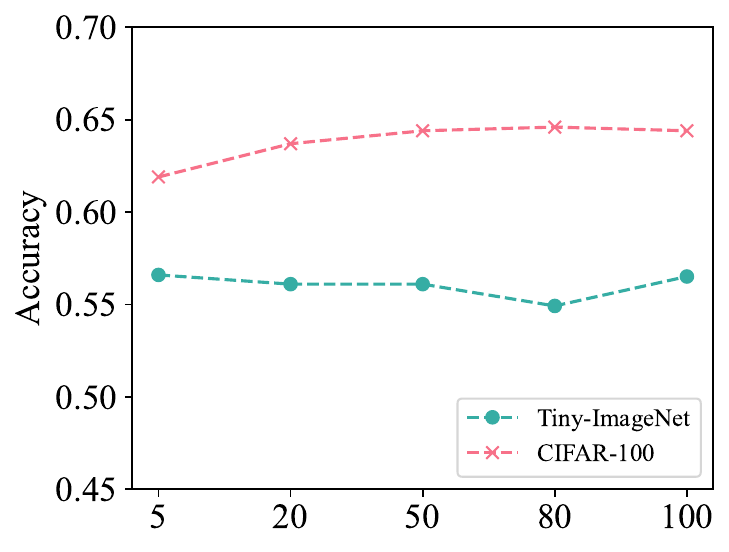}
\end{minipage}
}
\qquad
\subfigure[Influence of $\alpha$ .]{   
\begin{minipage}{4.7cm}
\flushleft     
\includegraphics[scale=0.35]{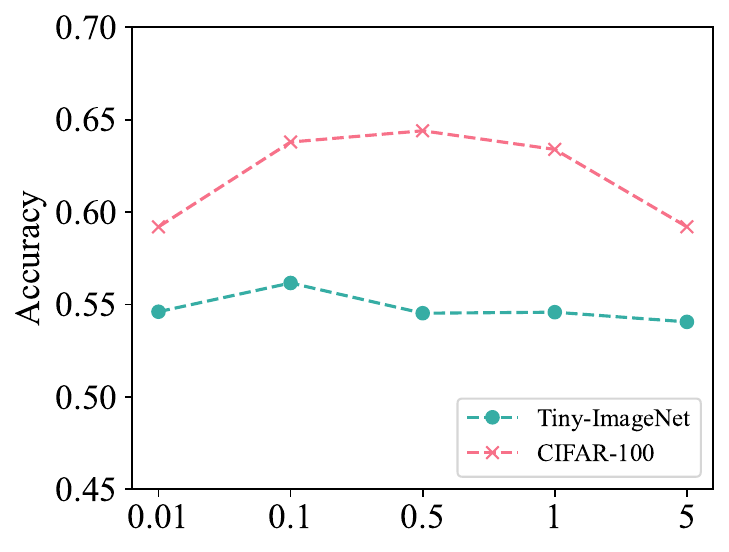}  
\end{minipage}
}

\subfigure[Influence of $\beta$ .]{ 
\begin{minipage}{4.7cm}
\flushleft     
\includegraphics[scale=0.35]{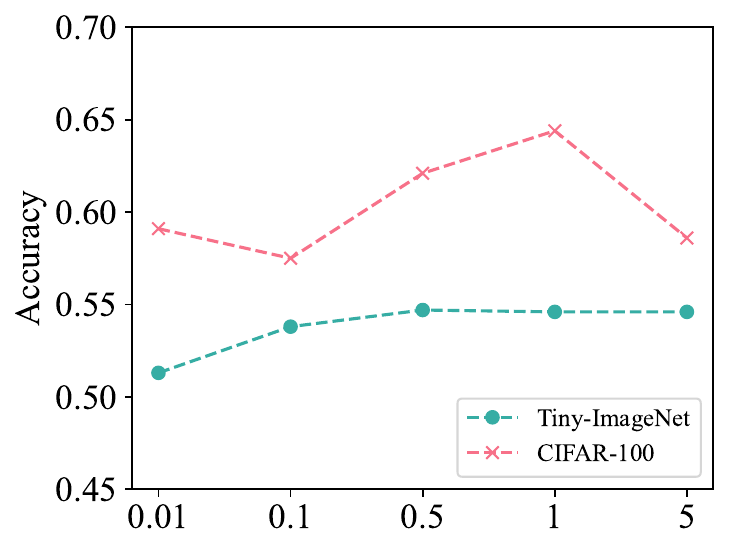}
\end{minipage}
}
\qquad
\subfigure[Influence of $\gamma$.]{ 
\begin{minipage}{4.7cm}
\flushleft    
\includegraphics[scale=0.35]{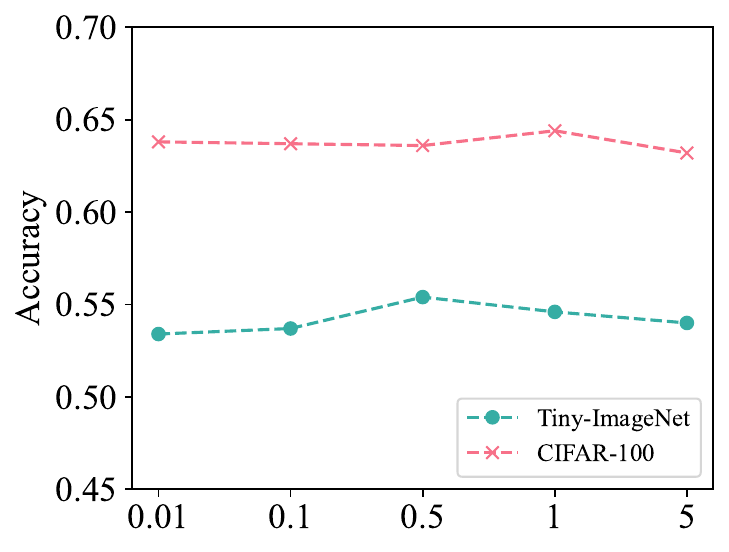}
\end{minipage}
}
\qquad
\subfigure[Influence of $\delta$.]{ 
\begin{minipage}{4.7cm}
\flushleft     
\includegraphics[scale=0.35]{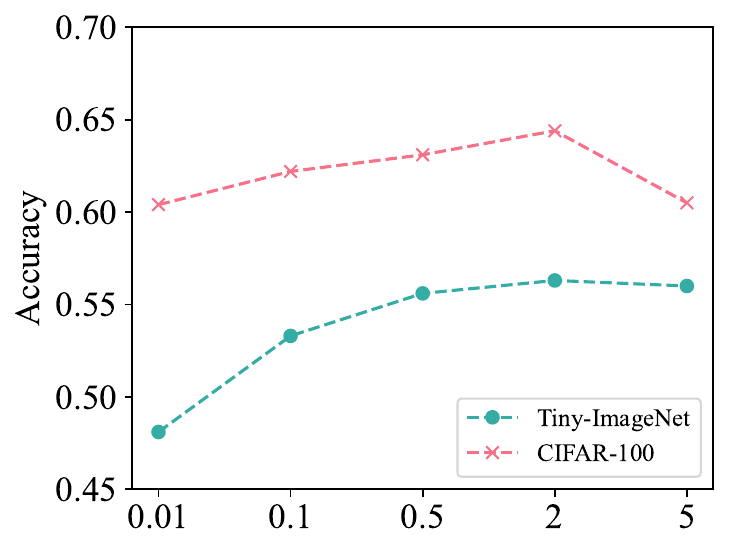}
\end{minipage}
}


\caption{Parameter Sensitivity Analysis. Backbone: ViT-B/16. Metric: ACC.}    
\label{sensitivit}    
\end{figure*}


\subsection{Parameter Sensitivity Analysis}

\subsubsection{Parameters in Image-Text Pairs Generation Stage}
We primarily focused on the ablation of $k_1$ and $k_2$, which significantly impact the quality of the generated text features. We can see from Fig.~\ref{sensitivit}~(a) and (b) that setting $k_1$ and $k_2$ too large or small will reduce the clustering performance. Intuitively, when $k_1$ or $k_2$ is too small, there are too few noun features used to construct the text feature, resulting in less rich semantic information in the generated features. Conversely, if $k_1$ or $k_2$ is overlarge, it will introduce too many irrelevant nouns, thus lowering the quality of the generated text feature. We adopted a configuration that performs well across all datasets, with  $k_1=200$ and  $k_2=50$, ensuring the presence of both rich nouns and high-quality text features.

\subsubsection{Loss Weights in Cross-Modal Semantic Consistency Learning Stage}
We also performed sensitivity analysis on the loss weights $\alpha, \beta, \gamma$ and $\delta$, which are involved in cross-modal semantic consistency learning loss, to prove the robustness of our method, as shown in Fig.~\ref{sensitivit}~(c), (d), (e), and (f), respectively. 
It can be found that the performance of our proposed method is stable on Tiny-ImageNet when all loss weights vary from 0.1 to 5. On the CIFAR-100, our method is also insensitive to all weights over specific intervals, such as $\alpha$ varying between $\{0.1, 0.5, 1\}$.

\subsection{Analysis of the Self-Enhanced Efficient Fine-tuning}
To better understand our proposed self-enhanced efficient fine-tuning strategy, we demonstrated the impact of different LoRA adapter settings on the clustering results and reveal the beneficial effects of different losses for self-enhanced learning on the clustering performance.

\begin{table}[!t]
\caption{
Ablation studies on different adapter settings. Dataset: CIFAR-100. Backbone: ViT-B/16. Metric: NMI.}
\label{ablationAdapter}\centering
\begin{threeparttable}
\resizebox{0.65\linewidth}{!}{\begin{tabular}{cccc}
\hline
Parallel     & FFN          & ReLU         & CIFAR-100 \\ \hline
$\checkmark$ &              &              & 0.732     \\
$\checkmark$ &              & $\checkmark$ & 0.724     \\
$\checkmark$ & $\checkmark$ &              & 0.717     \\
$\checkmark$ & $\checkmark$ & $\checkmark$ & 0.730     \\
             &              &              & 0.723     \\
             &              & $\checkmark$ & 0.714     \\
             & $\checkmark$ &              & 0.717     \\
             & $\checkmark$ & $\checkmark$ & 0.717     \\ \hline
\end{tabular}}

        
    \end{threeparttable}
\end{table}

\subsubsection{Adapter Settings}
In ~\cite{he2021towards}, a comprehensive comparison of various adapter settings revealed that the parallel adapter outperforms the serial adapter, and the adapter applied to FFN is more effective than that applied to $Q$ and $V$. Inspired by this work, we evaluated three LoRA adapter scenarios in the unsupervised clustering setting: whether the adapter is parallel or serial, whether the adapter is applied to FFN or to Q and V, and whether ReLU is added to LoRA. Comparison results are shown in Tab.~\ref{ablationAdapter}. It can be observed that in the setting of deep clustering, the parallel LoRA adapter still outperforms the serial adapter, but applying an adapter to the FFN and adding ReLU have a negative impact on performance.

\subsubsection{Self-Enhanced Learning Losses}\label{sell}
Tab.~\ref{ablationfintune} compares different loss functions for the self-enhancement stage. As hypothesized in our method section, continuing to fine-tune with the original alignment loss ($L_{align}$) causes a performance collapse (24.7\% ACC). This is because the vision encoder's features drift away from the static, pre-computed text features. In contrast, using a self-supervisory signal based on pseudo-labels is highly effective. Both FixMatch~\cite{sohn2020fixmatch} and our proposed confidence-weighted loss ($L_{self}$) yield strong results. Our $L_{self}$ performs best, demonstrating the benefit of dynamically weighting samples based on prediction confidence.

\begin{table}[!t]
\caption{Effectiveness of self-enhanced learning loss. Dataset: CIFAR-100. Backbone: ViT-B/16. Metric: ACC.}
\label{ablationfintune}\centering
\resizebox{0.7\linewidth}{!}{\begin{tabular}{cc}
\hline
Self-Enhanced Learning Loss    & CIFAR-100 \\ \hline
Original Loss  & 0.247     \\
FixMatch Loss & 0.704     \\
Self-enhanced Loss & 0.715     \\ \hline
\end{tabular}}
\end{table}

\section{Conclusion}
\label{sec:conclusion}
In this paper, we introduced SEIC, a self-enhanced framework that significantly improves deep image clustering by addressing the feature mismatch problem in frozen pre-trained models. Our two-stage "align, then enhance" approach first establishes a strong baseline by aligning clustering heads with CLIP's semantics through multi-level cross-modal consistency. This alignment is strengthened by our novel probability-weighted cluster center generation and a dynamic balancing regularizer. Subsequently, the model feeds back its own high-confidence predictions to efficiently fine-tune the vision encoder, adapting it to the specific task. Extensive experiments validated that SEIC sets a new state-of-the-art on six benchmarks. As a limitation, our reliance on explicit textual semantics may hinder performance in language-limited domains like medical imaging. Future work could explore generating pseudo-textual cues or developing non-language-based adaptation methods.
\clearpage

\bibliography{aaai2026}


\clearpage
\appendix
\twocolumn[
  \begin{center}
    \vspace{-1em} 
    {\Large\bfseries Supplementary Material} \\
    \vspace{1.2em}
    {\large\bfseries for "Self-Enhanced Image Clustering with Cross-Modal Semantic Consistency"}
    \vspace{1.5em}
  \end{center}
]
\setcounter{section}{0}
\section{Experimental Details Settings}
\label{sec:DetailsSettings}
\subsection{Dataset}
\label{sec:Dataset}

We evaluated the performance of our model by six widely recognized datasets, namely CIFAR-10, CIFAR-100, STL-10, ImageNet-10, ImageNet-dogs, and Tiny-ImageNet. Regardless of the initial dimensions of the images in these datasets, we uniformly resized them to $224\times 224$ pixels during training.

\begin{itemize}

\item Both CIFAR-10 and CIFAR-100 datasets comprise 60,000 images~\cite{krizhevsky2009learning}.   CIFAR-10 is composed of 10 categories, whereas CIFAR-100 contains 20 superclasses, with each superclass encompassing 5 categories.   In our training procedure, we exclusively utilized the 20 superclasses.

\item STL-10 comprises 10 categories of images, encompassing a substantial number of unlabeled images and smaller images~\cite{coates2011analysis}.    For our analysis, we exclusively employ the 13,000 labeled images from this dataset.

\item ImageNet-10 and ImageNet-dogs are subsets of the larger ImageNet~\cite{deng2009imagenet} collection. ImageNet-10 is constructed from a random selection of 10 categories within ImageNet, comprising a total of 13,000 images.    ImageNet-dogs, on the other hand, consists of 15 randomly chosen categories of dog images from ImageNet, totaling 19,500 images.

\item Tiny-ImageNet is a more extensive subset of ImageNet~\cite{le2015tiny}, encompassing 200 categories and a total of 100,000 images.


\end{itemize}

\subsection{Models}
We primarily conducted experiments using the pre-trained CLIP models ViT-B/32 and ViT-B/16, which consist of a vision encoder and a text encoder.   Each encoder is accompanied by a projection head and a clustering head.   When the encoder transforms features into a $D$-dimensional space, and the images have $K$ classes, both the vision projection head and the text projection head are $D\times D$ linear layers.   Meanwhile, the vision clustering head and the text clustering head are $D\times K$ linear layers.   In the case of ViT-B/32 and ViT-B/16, $D$ is set to 512.

\subsection{Training Settings}
In the image-text pairs generation stage, we set $k_1$ and $k_2$ to 200 and 50. During the cross-modal semantic consistency learning phase,  we employed the Adam optimizer with a learning rate of 0.005 for 200 epochs and a batch size of 1024. Our hyperparameters are set as follows: $\alpha=0.5$, $\beta=1$, $\gamma=1$, and $\delta=2$. In the self-enhanced efficient fine-tuning process, training is focused solely on the vision encoder. We introduce parallel LoRA layers to the $Q$ and $V$ components. If the dimensions of $Q$ or $V$ are $D$, the LoRA structure comprises just two linear layers: one is a $D\times r$ linear layer, and the other is a $r\times D$ linear layer. For this experiment, we set the LoRA layer's $r$ value to 128. The AdamW optimizer is adopted with a learning rate of 0.00005 for 40 epochs and a batch size of 128. 

\subsection{Augmentation}

In the cross-modal semantic consistency learning phase, we employed only basic image augmentation techniques, including Resize, CenterCrop, and Normalize, to preserve the original and authentic information of the images as much as possible.    During self-enhanced efficient fine-tuning, we utilized more complex image augmentation methods to allow the model to learn richer information, such as RandomResizedCrop, RandomHorizontalFlip, ColorJitter, and RandomGrayscale. During the evaluation, we used only Resize and CenterCrop.

\section{Pseudo Code}

We summarize the whole training process in Algorithm 1.

\section{Standard deviation}
With ViT-B/16, clustering accuracy is 0.708$\pm$ 0.010 and  0.898$\pm$ 0.014 on CIFAR-100 and ImageNet-Dogs.
 

\begin{algorithm}[t]
 \caption{ Training Algorithm}\label{alg:alg1}
 
 \begin{algorithmic}
 \STATE
  \STATE\textbf{Input:}\ Images $\mathcal{I}=\{I_i\}_{i=1}^N$, number of clusters $K$, a pre-trained CLIP vision encoder $f_v$ and text encoder $f_t$.
 \STATE \textbf{Output:}\ Clustering Results $\mathcal{C}$.
  \STATE Initializing head components $g_v$ , $g_t$ ,  $c_v$ and $c_t$.
  \STATE \textbf{Stage1}: Image-Text Pairs Generation
  \STATE Calculating noun features by WordNet and constructing text features $t_i$ for every image $I_i$ by Eq.~(1);
  \STATE \textbf{Stage2}: Cross-Model Semantic Consistency Learning
\FORALL {each batch}
  \STATE Calculating cross-modal semantic consistency learning loss by Eq.~(3), (5), (8), and (11);
  \STATE Update parameters of $g_v, g_t, c_v, $ and $c_t$ by Eq.~(11);
\ENDFOR
 \STATE \textbf{Stage3}: Self-Enhanced Efficient Fine-tuning
 \STATE Initializing LoRA adapters.
 \FORALL {each batch}
  \STATE Calculating the weights $w_i$ of images;
  \STATE Update parameters of $f_v, g_v, $ and $c_v$ by Eq.~(13).
 \ENDFOR

 \STATE Generating results $\mathcal{C}$ by $\arg \max (c_v(g_v(f_v(I_i))))$.
\end{algorithmic}
\end{algorithm}

\begin{table*}[!ht]
\caption{Performance of self-enhanced efficient fine-tuning. $\dagger$ indicates that the self-enhanced efficient fine-tuning strategy is not used. Backbone: ViT-B/16.}
\label{moredataset}
\resizebox{\linewidth}{!}{
\begin{tabular}{cccccccccccccccccccc}
\hline
\multicolumn{1}{c}{Dataset}            & \multicolumn{3}{c}{CIFAR-10}                                                & \multicolumn{3}{c}{CIFAR-100}                                               & \multicolumn{3}{c}{STL-10}                                                  & \multicolumn{3}{c}{ImgaNet-10}                                              & \multicolumn{3}{c}{ImageNet-Dogs}                                           & \multicolumn{3}{c}{Tiny-ImageNet}                                           \\ \hline
\multicolumn{1}{c}{Method}   & \multicolumn{1}{c}{NMI} & \multicolumn{1}{c}{ACC} & \multicolumn{1}{c}{ARI} & \multicolumn{1}{c}{NMI} & \multicolumn{1}{c}{ACC} & \multicolumn{1}{c}{ARI} & \multicolumn{1}{c}{NMI} & \multicolumn{1}{c}{ACC} & \multicolumn{1}{c}{ARI} & \multicolumn{1}{c}{NMI} & \multicolumn{1}{c}{ACC} & \multicolumn{1}{c}{ARI} & \multicolumn{1}{c}{NMI} & \multicolumn{1}{c}{ACC} & \multicolumn{1}{c}{ARI} & \multicolumn{1}{c}{NMI} & \multicolumn{1}{c}{ACC} & \multicolumn{1}{c}{ARI} \\ \hline
~SEIC $^\dagger$  & 0.839 & 0.925 & 0.843 & 0.618 & 0.644 & 0.489 & 0.966 & 0.987 & 0.972 & 0.984 & 0.994 & 0.988 & 0.790 & 0.869 & 0.763 & 0.649 & 0.561 & 0.405\\

SEIC                        & 0.945                   &    0.978                & 0.952                   & 0.732                  & 0.715                   & 0.598                   & 0.980                    & 0.993                   & 0.985                   & 0.992                  & 0.997                  & 0.994                  & 0.886               & 0.910                    & 0.858                   & 0.730                    & 0.644                 & 0.523                 \\ \hline
\end{tabular}}
\end{table*}

\begin{figure*}[!ht]
    \subfigure[CLIP vision encoder]{
        \begin{minipage}{0.3\linewidth}
        \centering    
        \includegraphics[width=1.0\textwidth]{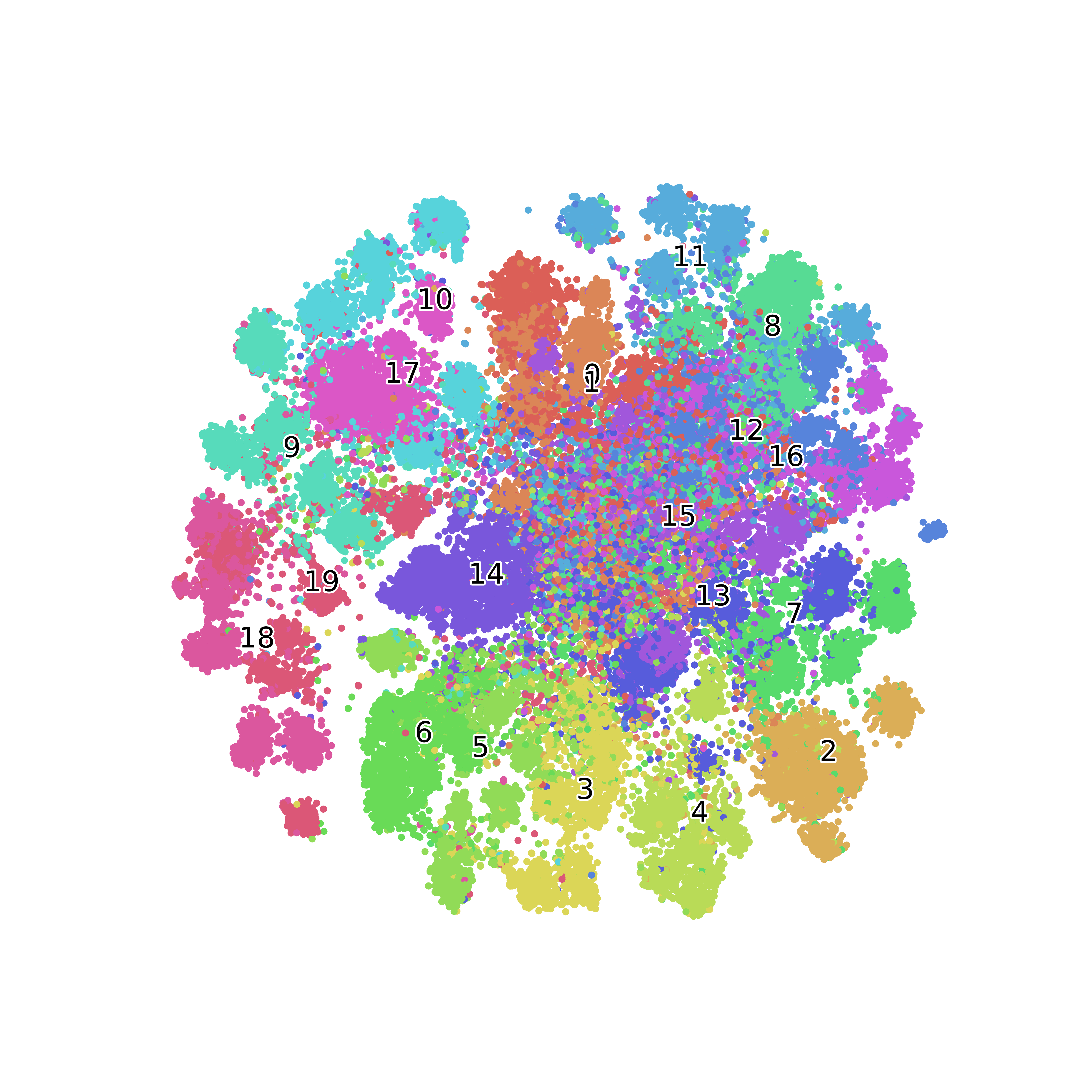}  
        \end{minipage}
    }
    \,
    \subfigure[Cross-modal semantic consistency learning]{
        \begin{minipage}{0.3\linewidth}
        \centering    
        \includegraphics[width=1.0\textwidth]{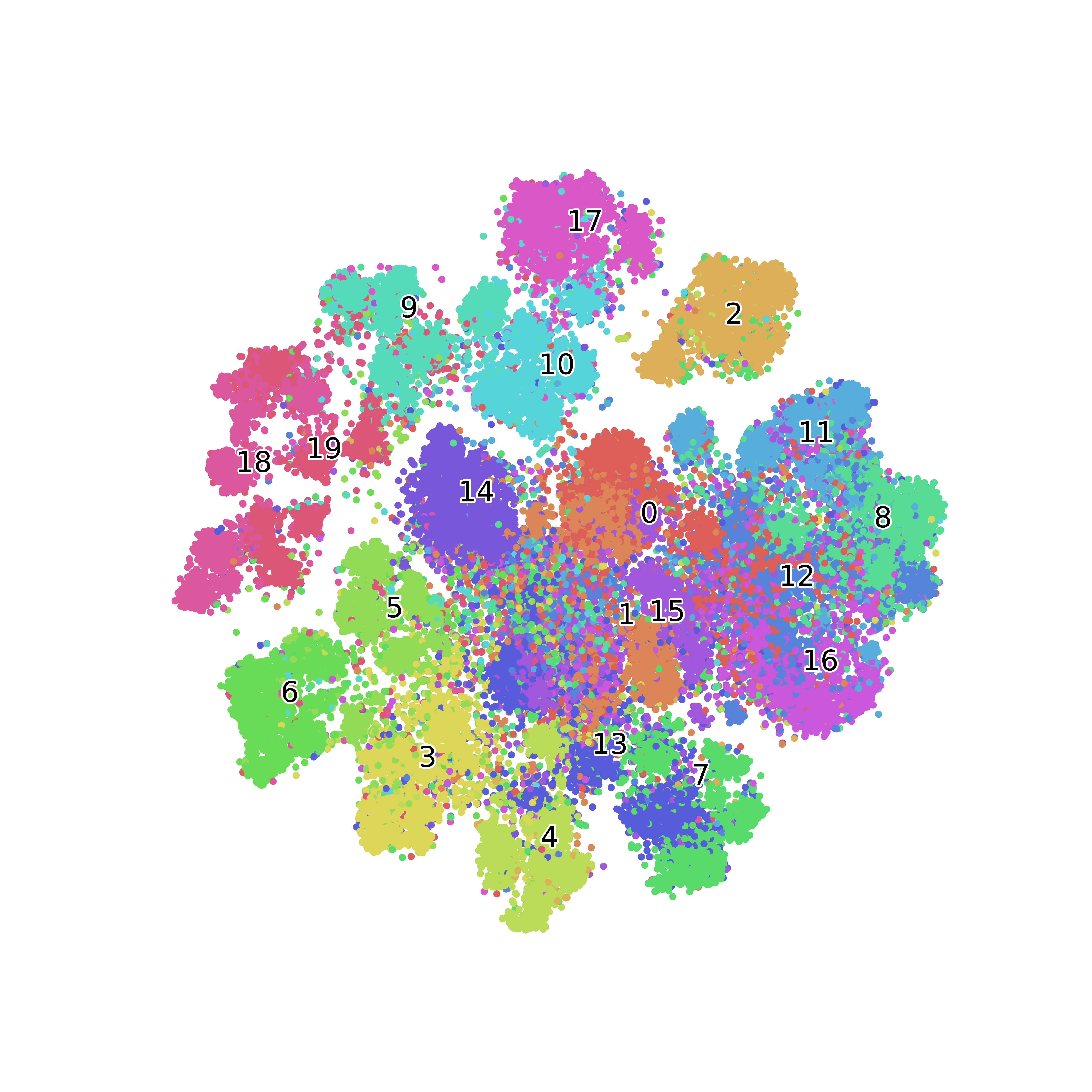}  
        \end{minipage}
    }
    \,
    \subfigure[Self-enhanced efficient fine-tuning ]{
        \begin{minipage}{0.3\linewidth}
        \centering    
        \includegraphics[width=1.0\textwidth]{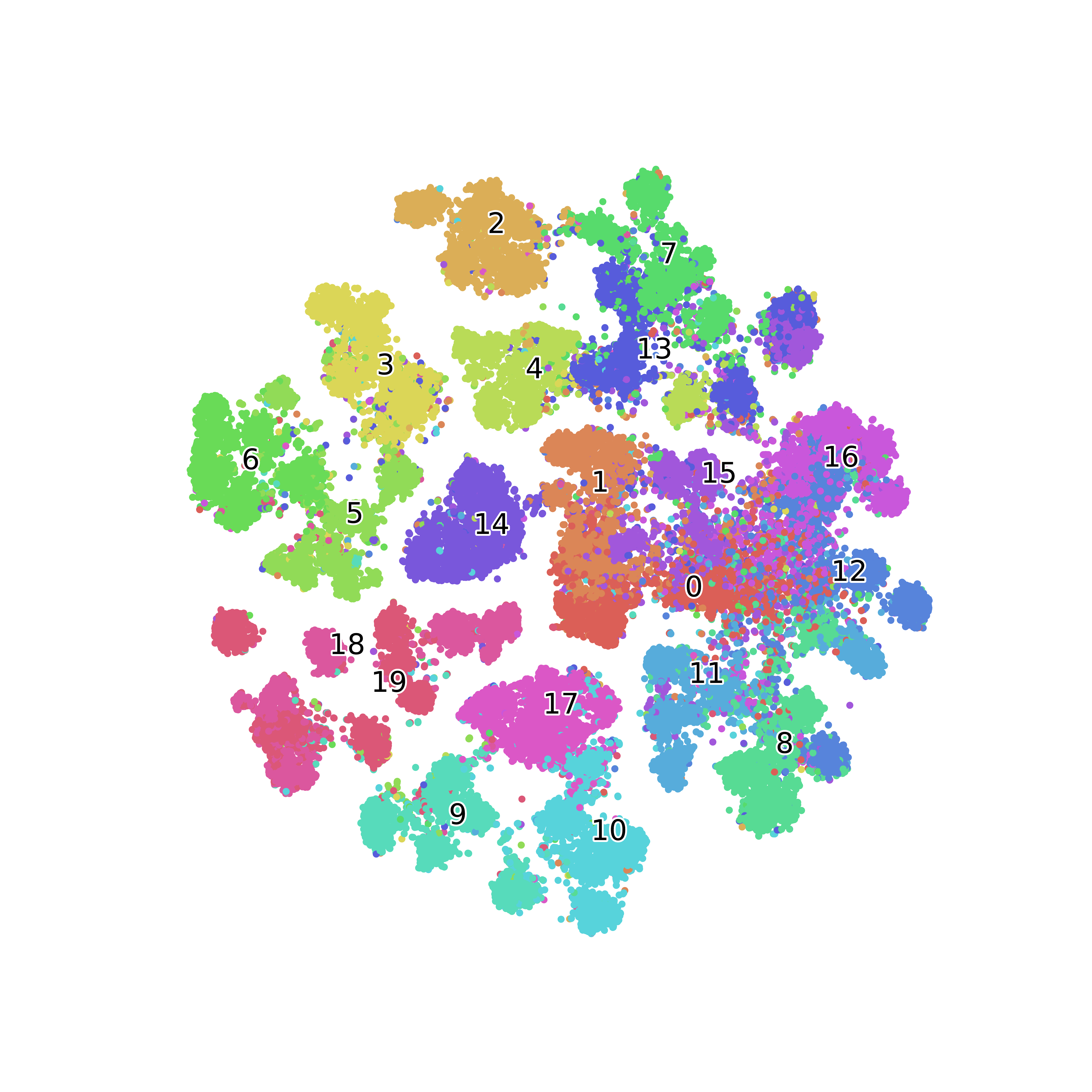}  
        \end{minipage}
    }
    \caption{Visualization of feature representations with t-SNE on CIFAR-100. Backbone: ViT-B/16 } 
    \label{fig:tsne}
\end{figure*}

\section{ Performance of Self-enhanced Efficient Fine-tuning}
We have illustrated the effects of self-enhanced efficient fine-tuning on two datasets. We extend our demonstration to include additional datasets in Tab.~\ref{moredataset}. It can be observed that improvements have been achieved across all datasets.

\section{Results of Text features}
The text features generated by our method are not based on ground-truth annotations but are instead synthesized from visual information via WordNet. To validate their semantic relevance, we performed a direct clustering experiment on these generated text features alone. Using K-means on the ViT-B/32 text features yielded a clustering accuracy of 83.0\% on CIFAR-10 and 45.4\% on CIFAR-100. These strong results confirm that our generated text features effectively capture meaningful semantic information, making them a valuable source for cross-modal consistency learning.

\section{Visualization of Features}
In order to intuitively understand the role of the different training stages, we visualized the image features generated at each stage on the CIFAR-100 dataset by t-SNE. From Fig.~\ref{fig:tsne}, we can observe that the clusters generated by CLIP mostly gather together without clear boundaries. When we perform the cross-modal semantic consistency learning, more and more clusters tend to be separated from each other. This trend reaches its peak when we adopt the self-enhanced efficient fine-tuning strategy. The whole process fully demonstrates how our proposed framework learns the discriminative features.

\section{Case Study}

\begin{figure*}[!t]
\centering
\includegraphics[width=0.95\textwidth]{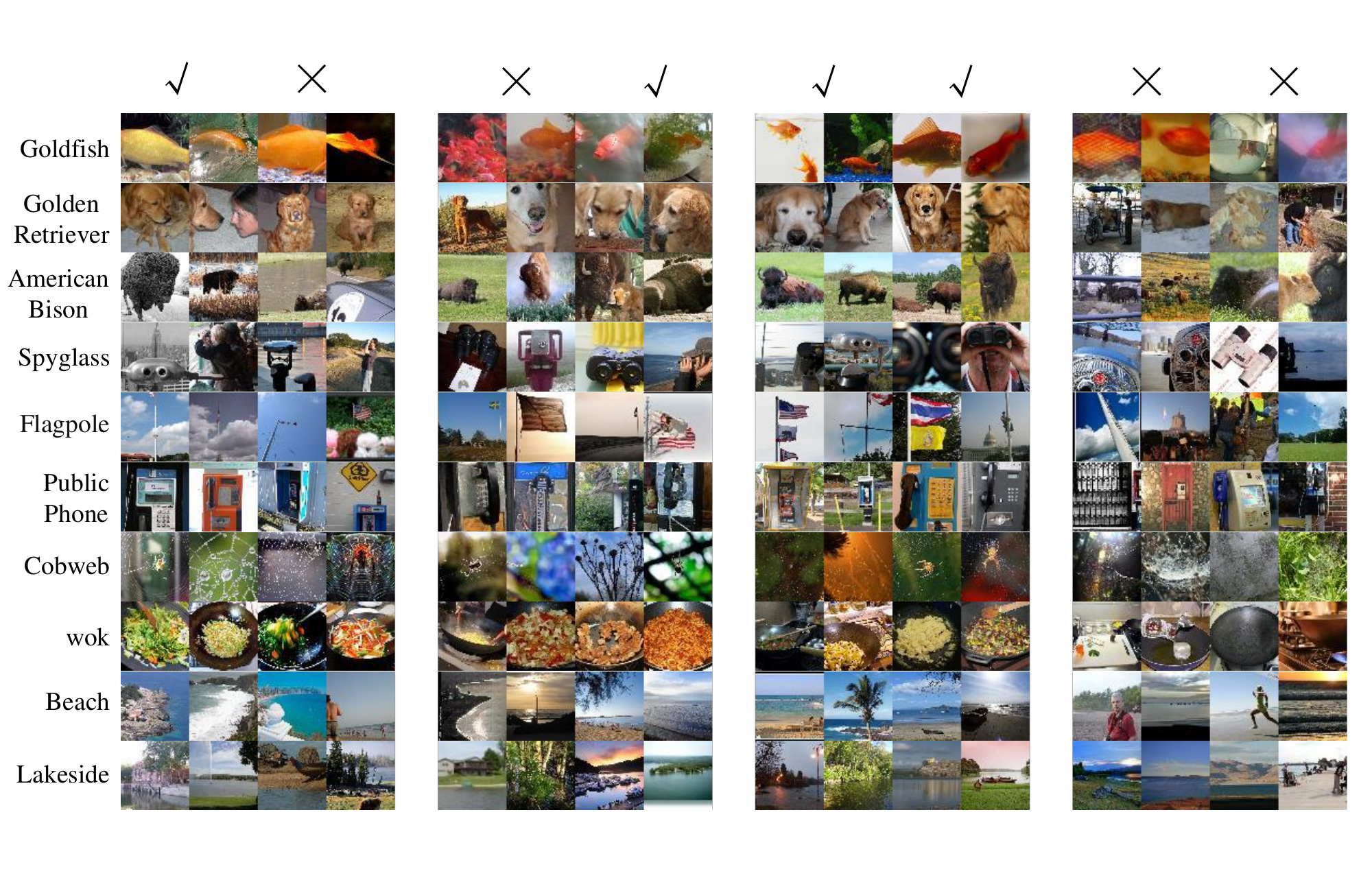}
\caption{A case study involving 10 categories from Tiny-ImageNet is presented. The figure illustrates four columns of images, with each column containing four images. Above each column, two symbols, $\checkmark$ or $\times$, are annotated. The first symbol in each column indicates whether the clustering was correct during the Cross-Modal Semantic Consistency Learning phase, while the second symbol denotes the correctness of clustering after the Self-Enhanced Efficient Fine-tuning phase. Taking the first column as an example, the images in this column were accurately clustered during the Cross-Modal Semantic Consistency Learning phase but were incorrectly clustered following the Self-Enhanced Efficient Fine-tuning phase, and so forth. Overall, each phase demonstrates eight images that were correctly classified and eight that were misclassified.}
\label{case_seic}
\end{figure*}

To intuitively demonstrate the final clustering outcomes, Fig.~\ref{case_seic} presents a selection of clustering results on the Tiny-ImageNet dataset. Given that the SEIC method employs a two-phase strategy, although the self-enhanced efficient fine-tuning phase significantly enhances clustering performance, the improvement is not solely attributable to incremental contributions. Specifically, there may be instances where images are correctly classified prior to fine-tuning but are misclassified thereafter.
The evaluation of image clustering tasks carries a degree of subjectivity. Taking the 'flagpole' category as an example, post self-enhanced efficient fine-tuning, the model has increasingly captured the 'flag' information, leading to the absence of flag elements in images that were misclassified at this stage. This indicates that the model has grasped sufficient similarity information, yet its classification criteria diverge from human expectations. Similarly, the 'wok' category reflects this issue: In images correctly classified across both phases, the wok contains dishes, whereas misclassified images depict a clean wok.
Moreover, some misclassified images may be influenced by subject interference. For instance, in the 'beach' category, misclassified images include human figures; in the 'golden retriever' category, interference from people or vehicles leads to model misclassification.

\section{Discussion}
In this section, we discuss the differences between our method and previous works.

\subsection{Discussion about PCL and ProPos} 
The Cross-Modal Semantic Consistency Learning phase of our method, along with PCL and ProPos, generates cluster centers within a batch for cluster center level comparison.

The methodologies employed by PCL and ProPos for generating cluster centers are fraught with two principal issues.  Firstly, the accuracy of the generated cluster centers is suboptimal.  Secondly, the process of generating these cluster centers is bifurcated into two stages, necessitating the assistance of K-means, which renders the approach inefficient.

Whereas PCL and ProPos typically average samples within the same categories to generate cluster centers, our methodology employs a novel approach for the generation.
Specifically, we employ $L_{ass}$ at the clustering head to evaluate the distance of samples to the actual cluster centers.  By weighting samples within a batch based on their proximity to these centers, we facilitate the emergence of high-quality cluster centers. 

\subsection{Discussion on  TAC}
TAC generates positive samples by leveraging nearest neighbors and employs a training loss composed of two elements: one analogous to $L_{ass}$, and another that functions as a regularization loss, as in previous works like GCC and SCAN. Our methodology improves the contrastive layer by facilitating comparisons across three discrete levels: instance feature, cluster assignment, and cluster center. The regularization loss is improved to make the cluster assignments more uniform. Moreover, we have effectively self-enhanced the backbone network with LoRA without resorting to any labeled data.

\end{document}